\documentclass[conference]{IEEEtran}
\IEEEoverridecommandlockouts
\usepackage{cite}
\usepackage{url}
\usepackage{xfrac}
\usepackage{booktabs}
\usepackage{amsmath,amssymb,amsfonts}
\usepackage{algorithmic}
\usepackage{graphicx}
\usepackage{textcomp}
\usepackage{xcolor}
\def\BibTeX{{\rm B\kern-.05em{\sc i\kern-.025em b}\kern-.08em
    T\kern-.1667em\lower.7ex\hbox{E}\kern-.125emX}}
\begin{document}

\title{Glass Surface Segmentation with an RGB-D Camera via Weighted Feature Fusion for Service Robots\\
\thanks{\textit{\underline{Citation:}} H. Lin, Z. Zhu, T. Wang, A. Ioannou, and Y. Huang. ``Glass Surface Segmentation with an RGB-D Camera via Weighted Feature Fusion for Service Robots," 2025 6th International Conference on Computer Vision, Image and Deep Learning (CVIDL), pp. 517-524, IEEE, 2025.}
}

\author{\IEEEauthorblockN{Henghong Lin$^{1}$, Zihan Zhu$^{1}$, Tao Wang$^{1,\ast}$, Anastasia Ioannou$^{2}$, Yuanshui Huang$^{3}$}
\IEEEauthorblockA{$^{1}$Fujian Provincial Key Laboratory of Information Processing and Intelligent Control, Minjiang University, Fuzhou, China. \\
$^{2}$Department of Computer Science and Engineering, European University Cyprus, Nicosia, Cyprus.\\
$^{3}$Fujian Hantewin Intelligent Technology Co., Ltd., Fuzhou, China.}
$^{*}$Corresponding author: Tao Wang, E-mail: twang@mju.edu.cn.}

\maketitle

\begin{abstract}
We address the problem of glass surface segmentation with an RGB-D camera, with a focus on effectively fusing RGB and depth information. To this end, we propose a Weighted Feature Fusion (WFF) module that dynamically and adaptively combines RGB and depth features to tackle issues such as transparency, reflections, and occlusions. This module can be seamlessly integrated with various deep neural network backbones as a plug-and-play solution. Additionally, we introduce the MJU-Glass dataset, a comprehensive RGB-D dataset collected by a service robot navigating real-world environments, providing a valuable benchmark for evaluating segmentation models. Experimental results show significant improvements in segmentation accuracy and robustness, with the WFF module enhancing performance in both mean Intersection over Union (mIoU) and boundary IoU (bIoU), achieving a 7.49\% improvement in bIoU when integrated with PSPNet. The proposed module and dataset provide a robust framework for advancing glass surface segmentation in robotics and reducing the risk of collisions with glass objects.
\end{abstract}


\section{Introduction}

\begin{figure}[t!]
\centerline{\includegraphics[width=1\linewidth]{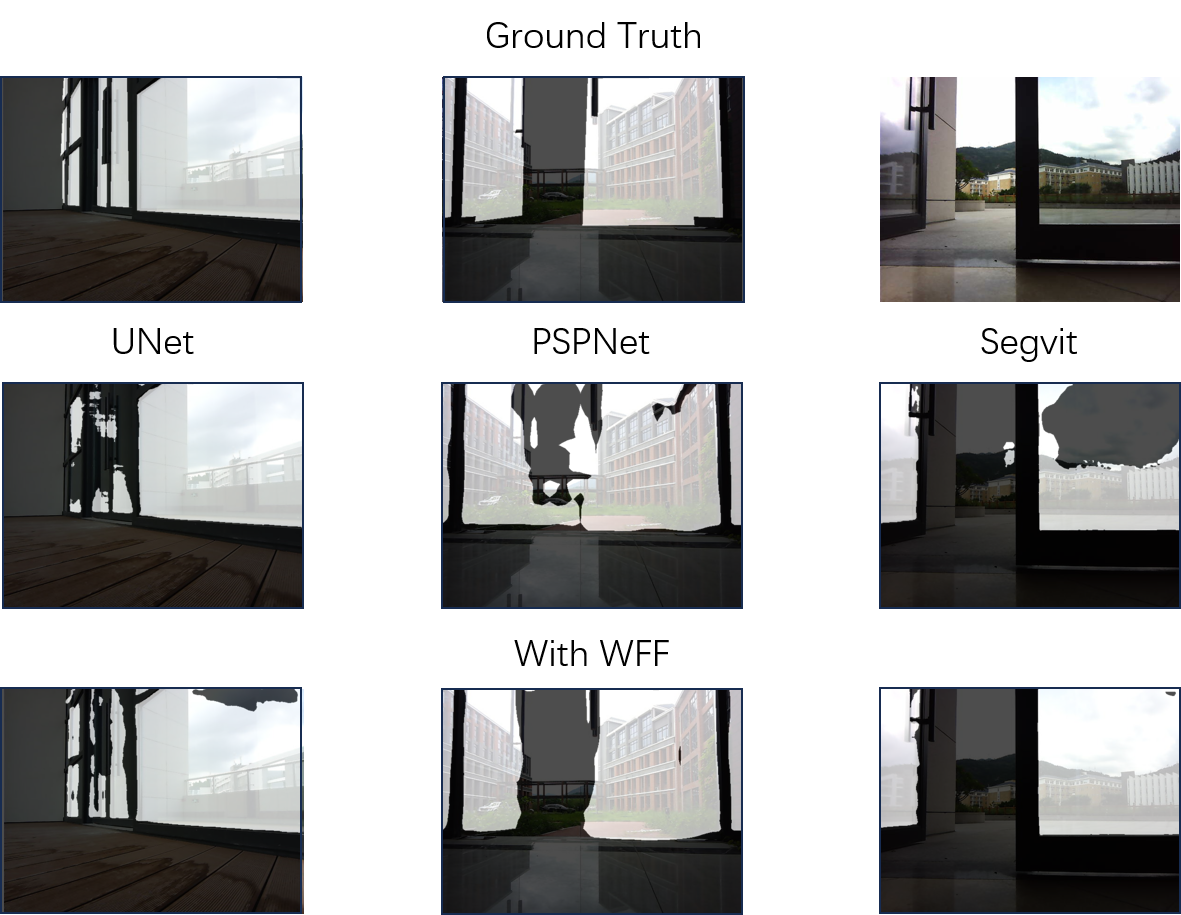}}
\caption{Examples of glass surface segmentation for service robots. The first row shows the input RGB image with the ground-truth glass surface segmentation overlaid. The second row shows results obtained with three representative methods: UNet~\cite{Ronneberger2015UNetCN}, PSPNet~\cite{Zhao2016PyramidSP} and SegViT~\cite{Zhang2022SegViTSS} using RGB images only. The third row shows results obtained with these methods using RGB-D images with Weighted Feature Fusion (WFF) proposed in this work. The results clearly demonstrate the superiority of RGB-D glass surface segmentation with WFF.}
\label{fig1}
\vspace{-5mm}
\end{figure}

With the rapid advancement of computer science, intelligent systems have become integral to daily human life. In this context, robotic systems are increasingly expected to perceive complex environments, accurately localizing objects and surfaces, estimate their spatial properties, and interact with them safely and effectively. A key capability enabling such perception is semantic segmentation, a fundamental task in computer vision and service robotics that involves assigning a category label to each pixel in an image. A wide range of methods have been proposed for semantic segmentation, with most relying on CNNs or ViTs to effectively capture both local and global contextual cues. Examples include FCN~\cite{Shelhamer2014FullyCN}, DeepLabv3 series~\cite{Chen2017RethinkingAC}, PSPNet~\cite{Zhao2016PyramidSP}, SETR~\cite{Zheng2020RethinkingSS} and SegViT~\cite{Zhang2022SegViTSS}. Although substantial progress has been achieved in generic object segmentation, segmenting semi-transparent objects such as glass surfaces remains a major challenge due to their inherent transparency, specular highlights, and complex interactions with surrounding objects. Additional difficulties arise from factors such as the lack of texture, background ambiguity, light refraction and reflection, and the absence of clear boundaries, all of which significantly complicate the segmentation process. Therefore, RGB images alone are typically insufficient for reliable glass surface segmentation, see second line of Fig.~\ref{fig1} for a few examples. Several methods have been proposed to address these challenges by incorporating additional input cues, such as infrared heat signatures~\cite{Huo2022GlassSW} and linear polarization information~\cite{Mei2022GlassSU}. Due to their cost-effectiveness and broad deployment in service robots, RGB-D cameras have emerged as a promising option among the various auxiliary sensing devices. In particular, glass and other reflective surfaces can be characterized by anomalous patterns in depth information with an RGB-D camera, such as missing depth values, distorted depth readings, and irregular depth discontinuities. As demonstrated in third line of Fig.~\ref{fig1}, these irregularities can, in turn, be exploited as image features to improve the accuracy of glass surface segmentation. Such capability is particularly important in the context of service robotics, where robots are required to operate safely and autonomously in complex environments. Many of these man-made environments, such as homes, offices, and public buildings, contain a variety of transparent or reflective surfaces, including glass doors, windows, tabletops, and partitions.

Previous works have explored leveraging RGB-D data for glass segmentation. Early works such as~\cite{Wang2012GlassOL} use appearance and depth features to train a local classifier combined with a Conditional Random Field (CRF) to encode the context correlations. Other works include~\cite{Huang2018GlassDA}  and~\cite{Zhao2023GlassDI} that either use an additional ultrasonic sensor or introduce the more restricted piecewise planar assumption perpendicular to the floor for glass surfaces. More recently and perhaps closer to our work, Zhu et al.~\cite{zhu2024mffnet} proposed a multimodal feature fusion network for RGB-D glass object detection. Although the fundamental idea is similar to ours, they focus on a limited number of glass objects on a robot manipulator station. In contrast, we focus on a much more challenging service robot perspective that could result in depth images with poorer quality due to a greater distance range and the presence of large glass surfaces. In this scenario, our method is able to adaptively select the weights for depth features to mitigate the depth image quality issue, which is a main strength of our method. In addition, Lin et al.~\cite{lin2025leveraging} proposed an RGB-D glass surface segmentation method based on cross-modal context mining and depth-missing aware attention. Unlike their method, our method is more simple and straightforward as it does not involve the more complex multi-stage context-wise and channel-wise attention, as well as the auxiliary input of a binary depth-missing map. In addition, while their method uses a curated ensemble dataset RGB-D GSD, which is more diverse, we create a real-world glass surface segmentation dataset that focuses on glass surface segmentation for service robots. Our goal is to provide a simple and generic plug-and-play module that can be used in conjunction with different neural network backbones for the safe navigation of domestic robots.

More concretely, we propose a novel approach for glass surface segmentation by integrating a specialized \textbf{Weighted Feature Fusion (WFF) module}, which enhances the adaptive fusion of RGB and depth data for better glass segmentation. The WFF module optimizes the ability of the model to handle reflections, transparency, and occlusions, ensuring that glass boundaries are better preserved and background interference is minimized. By effectively fusing multi-modal data, the WFF module enhances the segmentation accuracy of glass regions, even in complex environments with strong reflectivity and environmental clutter. In addition, we introduce the \textbf{MJU-Glass dataset}, a comprehensive dataset designed specifically for glass surface segmentation with service robots. Previous glass datasets such as RGB-D GSD~\cite{lin2025leveraging} were curated from major well-known datasets, which may not be directly applicable to tasks such as robot navigation. Furthermore, MJU-Glass includes images of glass surfaces with diverse transparency levels, reflective properties, and environmental disturbances, captured from a physical robot's perspective to reproduce realistic data collection scenarios. It provides a valuable benchmark for evaluating and comparing glass segmentation models in both practical and adverse conditions. Our main contributions are as follows:
\begin{itemize}
\item We propose a novel Weighted Feature Fusion (WFF) module for dynamic and adaptive fusion of RGB and depth features to improve glass surface segmentation, effectively handling transparency, reflections, and occlusions. In particular, our WFF module can be used in conjunction with different deep neural network backbones as a plug-and-play module to boost segmentation performance.
\item We create the MJU-Glass dataset, a comprehensive dataset for benchmarking with real-world glass surfaces featuring varying transparency and environmental interference. This provides a public platform for evaluating segmentation models. The RGB-D image pairs are collected from a physical robot navigating diverse environments from two distinct perspectives, capturing realistic operation scenarios, with the setup derived from an actual robotic platform used in commercial products developed by a robotics company.
\item We achieve substantial improvements in both segmentation accuracy and robustness, demonstrating the effectiveness of the proposed WFF module under realistic operational conditions. This is validated by performance gains in metrics such as mean Intersection over Union (mIoU) and boundary IoU (bIoU). For instance, integrating WFF with PSPNet yields a notable 7.49\% improvement in bIoU. The improved precision in boundary segmentation facilitates more accurate SLAM-based environmental modeling, reducing the likelihood of unnecessary collisions with glass objects.
\end{itemize}

\section{OUR APPROACH}
In this section, we introduce our model, with a particular focus on the Weighted Feature Fusion Module (WFF), which serves as the core of our design. The WFF leverages adaptive weights to dynamically balance and integrate RGB and depth features, improving the capability of the model to handle complex and diverse scenarios effectively.

\subsection{Model Architecture}
\begin{figure}[t!]
\centerline{\includegraphics[width=1\linewidth]{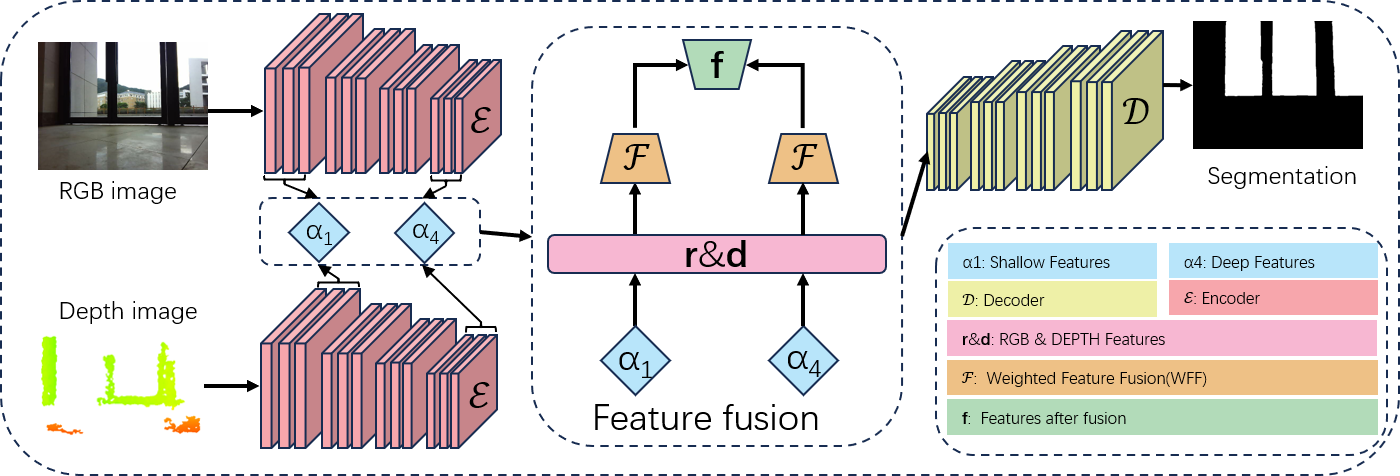}}
\caption{Overall model architecture. In this model, features of glass are fused across RGB and depth images to address the limitations of appearance features in RGB images for glass surface segmentation. The backbone extracts shallow and deep features from both inputs, which are fused in the WFF module and further processed in the decoder to generate the final segmentation result.}
\label{fig2}
\vspace{-2mm}
\end{figure}

We now describe the overall architecture of our model. In the following sections, we use the DeepLabv3+ framework with ResNet-50~\cite{He2015DeepRL} as an example for the backbone, but we note that the WFF modules proposed in this work are not limited to specific deep neural network architectures. The model takes RGB and depth images as input, which are passed through the backbone network to extract both shallow and deep features. The features are then fused together through the WFF module. The overall process is illustrated in Fig.~\ref{fig2}, which aims to enhance the performance of semantic segmentation tasks by incorporating depth information alongside conventional RGB data.

\textbf{Input Stage}: Both RGB and depth images are fed into the ResNet-50 backbone. The network extracts both shallow (low-level) and deep (high-level) features from each modality. Shallow features are captured by the initial convolutional layers and the first few residual blocks, which focus on low-level patterns such as edges and textures. Deep features, on the other hand, are extracted by the deeper residual blocks, capturing high-level semantic information and global context. This hierarchical feature extraction process leverages the strength of the neural network to represent diverse aspects of the input data effectively. These extracted features are then prepared for fusion in the WFF module.

\textbf{WFF Module}: In this stage, the RGB and Depth features are fused to generate fused shallow and deep features. This step ensures that the complementary information from the two modalities is effectively integrated, enhancing the ability of the model to identify glass features. Here, weighted fusion is adopted to adaptively adjust the weight values. In particular, it does not incur additional training computational overhead and can achieve better results compared to simple fusion.

\textbf{Decoder}: The fused shallow and deep features are concatenated to form the final feature map, which is used to predict the segmentation mask. This concatenation leverages both low-level and high-level information to achieve more accurate segmentation results.

\subsection{Overall Pipeline}
The proposed model takes as input an RGB image \(I\in R^{3\times H\times W}\) and a depth image \(D\in R^{H\times W}\) where \(H\) and \(W\) represent the height and width of the images, respectively. During training, for each image pair we are also supplied with a ground-truth mask \( M \in \{0, 1\}^{\mathrm{H \times W}} \), where 0 and 1 denote the background and the glass classes. The task is to perform semantic segmentation, classifying each pixel into one of two categories: background or glass. The objective is to accurately identify and segment glass surfaces within the input images.

Specifically, the model follows a structured pipeline, consisting of the following steps:

\textbf{Feature Extraction}: The RGB and depth images are first processed independently through the ResNet-50 backbone network to extract both low-level and high-level semantic features although other deep neural network backbones may also be used. ResNet-50, with its deep residual learning architecture, effectively handles the vanishing gradient problem, allowing it to capture more informative features across various scales. The backbone outputs multi-scale feature maps that capture both fine-grained spatial details for local edge detection and long-range contextual information for handling complex transparency effects in glass objects. Denote \(\alpha_1\) and \(\alpha_4\) as residual block 1 and residual block 4, and we denote the RGB feature map from \(\alpha_1\) as \(r_1\), depth feature map from \(\alpha_1\) as \(d_1\), and the RGB feature map from \(\alpha_4\) as \(r_4\) and depth feature map from \(\alpha_4\) as \(d_4\), respectively. We note that this is just an instantiation based on ResNet-50, and in the following discussion on feature fusion, we use notations with a lack of subscript \(r\) and \(d\) to refer to RGB and depth features for simplicity. Depth feature maps play a crucial role in distinguishing subtle edges that may be invisible in RGB images, particularly in challenging environments with transparent surfaces. The RGB modality provides rich color and texture information, while depth data complements this by offering critical spatial cues, such as distance and object boundaries, which are essential for accurately segmenting transparent surfaces like glass. The combination of both modalities enriches the feature representation, improving the ability of the model to handle a variety of complex scenes.

\textbf{Feature Fusion}: In order to efficiently integrate information from both RGB and depth models, we employ a channel-specific fusion mechanism. The intuition behind this mechanism is to allow the model to learn the optimal contribution of RGB and depth information for each channel, as different channels could capture different aspects of the appearance and depth features, which is crucial for the success of our method. Specifically, denote \(r\in R^{C\times h\times w}\) and \(d\in R^{C\times h\times w}\) as the RGB and depth features, respectively, and the feature fusion could be written as:
\begin{equation}
\scriptstyle
f = \mathrm{\Psi}_{RGB} \odot r + \mathrm{\Psi}_{DEPTH} \odot d,\ \text{where}\ \mathrm{\Psi}_{RGB}^c + \mathrm{\Psi}_{DEPTH}^c = 1,\ c = 1\ldots C,
\label{eqn1}
\end{equation}

\noindent where \(\mathrm{\Psi}_{\mathrm{RGB}}^c\) and \(\mathrm{\Psi}_{\mathrm{DEPTH}}^c\) are the pair of RGB and depth weights for the $c$-th channel. \(\mathrm{\Psi}_{RGB}\in R^C\) and \(\mathrm{\Psi}_{DEPTH}\in R^C\) a are the vectors of weights for all channels. In Eqn.~\ref{eqn1}, the spatial dimensions \(h\) and \(w\) of both \(\mathrm{\Psi}_{RGB}\) and \(\mathrm{\Psi}_{DEPTH}\) are first restored to match the dimensions of \(r\) and \(d\) before performing the Hadamard product \(\odot\) (i.e., element-wise multiplication). In other words, the weights are computed globally for each channel across the entire feature map. The output after fusion is f. See Sec.~\ref{sec:wff} for the detailed process of the fusion step.

By learning the optimal combination of RGB and depth through these weights, the model can adaptively allocate the most relevant information for each pixel. This adaptability is particularly critical for segmenting glass regions, where transparency, reflections, and objects behind the glass introduce significant visual challenges. The feature after fusion f is further refined through decoding layers discussed below, allowing the network to capture more discriminative information for the glass surface segmentation task.

\textbf{Decoding}: The fused features are passed to the decoder module, which up-samples the feature map to match the original input resolution. Multi-scale information is integrated by combining high-level semantic features with low-level spatial details. The decoder effectively recovers spatial resolution while preserving the global context learned during feature fusion, enabling the model to delineate glass boundaries more precisely.

\textbf{Classification}: Since the goal of the model is binary segmentation, i.e., the target classes are background and glass, for each pixel \(\left(i,j\right)\), the decoder outputs logits \(o_{i,j,0}\) and \(o_{i,j,1}\), correspond to class 0 (background) and 1 (glass), respectively. A softmax function is applied to the logits to compute the probability of each pixel belonging to class 0 or 1:
\begin{equation}
P\left(M_{ij}=k\right) = \frac{\exp\left(o_{i,j,k}\right)}{\exp\left(o_{i,j,0}\right) + \exp\left(o_{i,j,1}\right)}, \quad k \in \{0,1\},
\end{equation}

\noindent where \(M_{ij}\) denotes the element of \(M\) at \(\left(i,j\right)\). The softmax function normalizes the logits into probabilities, ensuring that each pixel is assigned a class with a probability that sums to 1, which is crucial for a binary segmentation task.

\textbf{Loss Function}: The model is trained using the binary cross-entropy loss, which compares the predicted probabilities with the ground-truth labels. The binary cross-entropy loss is computed for each pixel and then averaged across the entire image:
\begin{equation}
L_{BCE}=-\frac{1}{H\times W}\sum_{i=1}^{H}\sum_{j=1}^{W}{\sum_{k=0}^{1}M_{ij}log\left(P\left(M_{ij}=k\right)\right)},
\end{equation}

The binary cross-entropy loss is chosen for this task due to its simplicity and effectiveness in binary classification problems, especially when dealing with imbalanced datasets. This loss function computes the logarithmic difference between the predicted probability and the ground-truth label for each pixel, which is particularly well-suited for the task of segmenting two classes (glass and background). It is commonly used in pixel-wise classification tasks and allows the model to directly optimize for the pixel-level prediction.

\textbf{Prediction}: The final segmentation mask is obtained by assigning each pixel to the class with the highest predicted probability. To evaluate the model performance, we use metrics such as mean Intersection over Union (mIoU), Glass IoU, and Boundary IoU, which are chosen because they comprehensively assess the model accuracy in segmenting both the overall glass surfaces and its fine boundaries, which are particularly challenging due to the transparent nature of glass and the complex shapes due to occlusion and viewpoint variations.

This structured pipeline ensures that the complementary information from the RGB and depth modalities is effectively utilized, enhancing the ability of the model to segment glass objects accurately.

\subsection{Weighted Feature Fusion}
\label{sec:wff}
In this section, we present the detailed process of the feature fusion step. Our Weighted Feature Fusion (WFF) module is designed to efficiently fuse RGB and Depth features by dynamically adjusting their weights, as shown in Fig.~\ref{fig3}. Specifically, WFF involves the following sequential steps:

\begin{figure}[t!]
\centerline{\includegraphics[width=1\linewidth]{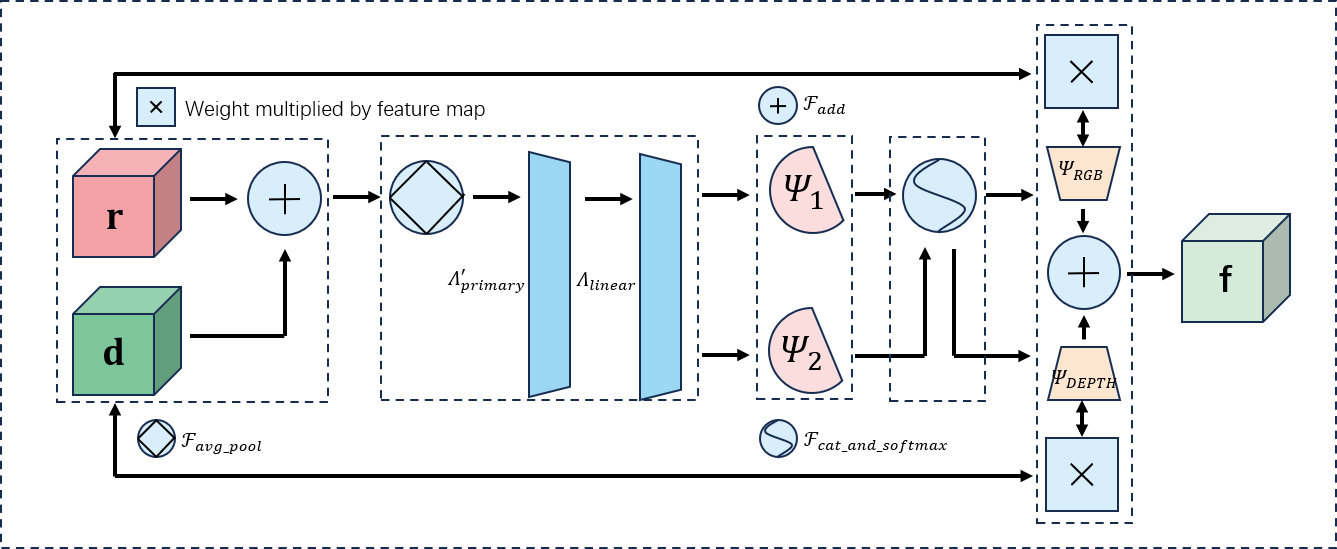}}
\caption{An illustration of the proposed Weighted Feature Fusion (WFF) module. The RGB and depth features \(r\) and \(d\) from the backbone network are first combined (\(\mathrm{\Lambda}_{primary}\)), and then the spatial resolution is reduced to \(1\times1\) with average pooling (\(\mathrm{\Lambda}_{primary}^\prime\)), in order to aggregate the global context of the image. The features are then separated through a two-layer perceptron, with a seperate second layer to produce two channel-specific weight vectors \(\mathrm{\Psi}_1\) and \(\mathrm{\Psi}_2\). These weights are then concatenated and normalized to obtain \(\mathrm{\Psi}_{RGB}\) and \(\mathrm{\Psi}_{DEPTH}\) with a softmax function to ensure a bounded representation with a probabilistic interpretation. Finally, the spatial dimensions of the channel weights are restored, and then applied to the input feature to obtain the output of WFF.}
\label{fig3}
\vspace{-2mm}
\end{figure}

\textbf{Feature Combination}: Initially, the \(c\)-th channel of the input feature maps \(r^c\) (RGB) and \(d^c\) (Depth) are combined through an element-wise addition operation \(F_{add}\), defined as:
\begin{equation}
\mathcal{F}_{add}:\mathrm{\Lambda}_{primary}=r^c+d^c,
\end{equation}

\noindent Here, element-wise addition is preferred over other operations like concatenation or channel-wise multiplication due to its simplicity and efficiency. By adding the features directly, the operation preserves both feature sets without introducing additional dimensionality, making it computationally efficient. Moreover, this approach retains the inherent structure of each feature map, ensuring that the combined features contribute to the model without unnecessary complexity.

\textbf{Resolution Reduction}: After feature combination, the feature map \(\mathrm{\Lambda}_{primary}\) undergoes an average pooling operation, \(F_{{avg\_pool}}\), reducing its resolution to \(1\times1\). This operation is applied to both high- and low-resolution features, aiming to aggregate global information and smooth feature variations across the entire spatial domain. By reducing the resolution, the module facilitates the computation of channel-wise weights in the subsequent fully connected layers. These weights allow the model to learn the optimal fusion strategy for RGB and depth features:
\begin{equation}
\mathcal{F}_{avg\_pool}:\mathrm{\Lambda}_{primary}^\prime=\frac{1}{h\times w}\sum_{i=1}^{h}\sum_{j=1}^{w}{\mathrm{\Lambda}_{primary}}_{i,j},
\end{equation}

\noindent where $h$ and $w$ are the height and width of \(\mathrm{\Lambda}_{primary}\), and \({\mathrm{\Lambda}_{primary}}_{i,j}\) is the \(\left(i,j\right)\)-th element of \(\mathrm{\Lambda}_{primary}\). While this resolution reduction is beneficial for integrating global context, it inevitably results in the loss of fine-grained spatial details, particularly in high-resolution features critical for precise boundary segmentation. To address this, the module utilizes the fused weights to reweight and project the original high-resolution features back into the spatial domain, ensuring that boundary details are preserved and enhanced in the final segmentation output. This strategy balances the trade-off between global context aggregation and the preservation of fine-grained boundary information.

\textbf{Preliminary Feature Separation}: The reduced feature map \(\mathrm{\Lambda}_{primary}^\prime\) is further processed by \(F_{{pre\_division}}\), which consist of two steps. The first step is a linear transformation by a fully-connected layer to obtain \(\mathrm{\Lambda}_{linear}\), which also halves the original feature dimension of \(\mathrm{\Lambda}_{primary}^\prime\). The second step involves two different fully-connected layers used to generate two different features (\(\mathrm{\Psi}_1\) and \(\mathrm{\Psi}_2\)). These two fully-connected layers apply respective linear transformations to \(\mathrm{\Lambda}_{linear}\), which preliminarily separates the features into \(\mathrm{\Psi}_1\) and \(\mathrm{\Psi}_2\). In the second step, the feature dimension is restored to the original feature dimension. The above steps can be written as follow:
\begin{equation}
\resizebox{0.9\linewidth}{!}{$
\mathcal{F}_{predivision}:\mathrm{\Lambda}_{linear} = {FC}_1\left(\mathrm{\Lambda}_{primary}^\prime\right),\ \mathrm{\Psi}_1 = {FC}_{21}\left(\mathrm{\Lambda}_{linear}\right),\ \mathrm{\Psi}_2 = {FC}_{22}\left(\mathrm{\Lambda}_{linear}\right)
$}
\end{equation}

\textbf{Weight Adjustment}: The preliminarily separated features \(\mathrm{\Psi}_1\) and \(\mathrm{\Psi}_2\) are concatenated along a new channel axis, forming a combined feature map that is then processed through the operation \(F_{\text{cat\_and\_softmax}}\). This step ensures that the combined feature map undergoes a softmax normalization along the concatenated channel axis, which guarantees that the values of the resulting fusion weights \(\mathrm{\Psi}_{RGB}\)  and \(\mathrm{\Psi}_{DEPTH}\) sums to $1$ for each channel. This produces a weight distribution for the RGB and depth features, representing their relative contributions to the final fused representation. 
\begin{equation}
\mathcal{F}_{\text{cat\_and\_softmax}}:
\left\{
\begin{aligned}
\Psi_{\text{RGB}} &= \frac{\exp(\Psi_1)}{\exp(\Psi_1) + \exp(\Psi_2)} \\
\Psi_{\text{DEPTH}} &= \frac{\exp(\Psi_2)}{\exp(\Psi_1) + \exp(\Psi_2)}
\end{aligned}
\right.
\end{equation}

Weight normalization using softmax is particularly important because it maps the output to the range $[0, 1]$ and ensures that the sum of the weights is equal to $1$, thus maintaining a balanced contribution from RGB and depth, whereas alternatives such as sigmoid treats each weight independently, which can result in an unbounded or disproportionate emphasis on features. This property makes softmax especially suitable for adaptive feature fusion, as it allows the model to dynamically adjust the weights of RGB and depth features based on the varying inputs. Unlike sigmoid, which can output values independent of each other (i.e., without ensuring a sum-to-$1$ constraint), or ReLU, which can produce unbounded outputs, softmax guarantees a normalized weight distribution. This makes it an ideal choice in multi-channel fusion scenarios where it is important to assign relative and normalized weights across channels, enhancing the flexibility and effectiveness of the feature fusion process.

\textbf{Weight Application and Fusion}: Finally, the spatial dimensions $h$ and $w$ are restored to the two channel-specific weight vectors \(\mathrm{\Psi}_{RGB}\) and \(\mathrm{\Psi}_{DEPTH}\). We note that these weights can be dynamically adjusted based on the input data. For instance, in well-lit scenes, the weight \(\mathrm{\Psi}_{RGB}\) will be more prominent, emphasizing RGB features that contain rich fine-grained spatial information. On the other hand, in low-visibility conditions (e.g., poor lighting), the weight \(\mathrm{\Psi}_{DEPTH}\) will play a larger role, allowing the model to rely more on the structural information provided by the depth map to maintain performance. This dynamic weighting allows the model to make optimal use of both types of feature under varying environmental conditions.

The fused feature map is computed with Eqn.~\ref{eqn1}, which is then passed to the decoder for further processing. This fusion operation effectively balances the fine-grained spatial information from the RGB features with the structural, depth-related information from the depth map, leading to a more robust and context-aware representation that adapts to the input scene. In our implementation, the decoder uses the features fused by the WFF module without employing skip connections. This is because, at the end of the WFF module, the original feature maps are used for the above weighted fusion operations, effectively preserving important spatial information. As a result, there is no need for skip connections from the backbone to the decoder to maintain spatial details.

\section{EXPERIMENTAL EVALUATION}
In this section, we present the experimental evaluation results of the proposed glass surface segmentation method with WFF. To thoroughly evaluate the performance of the proposed method on service robots, we create a large-scale RGB-D dataset MJU-Glass to verify the efficacy of WFF in different neural networks.
\subsection{Dataset}\label{AA}
\textbf{Overview and Dataset Collection}: Mei et al.~\cite{Mei2021DepthAwareMS} launched the RGBD-Mirror dataset by combining four mainstream datasets~(i.e. Matterport3D~\cite{Chang2017Matterport3DLF}, SUNRGBD~\cite{Song2015SUNRA}, ScanNet~\cite{Fan2019RethinkingRS} , and 2D3DS~\cite{Armeni2017Joint2D}), but these data were filtered from a number of datasets. Liang et al.~\cite{Liang2023MonocularDE} also constructed an RGB-D dataset with a special focus on glass walls only. In this work, we wanted to address the glass surface segmentation problem from the perspective of a service robot, which is less studied before. To this end, we collected a new RGB-D dataset, named MJU-Glass, using a wheeled robotic platform. The dataset is publicly available at \url{https://github.com/weduake/MJU_Glass}. In order to cover a wider view, we placed two RGB-D cameras at different heights and angles. The RGB-D cameras used were the Gemini Pro from Orbbec. The RGB images were captured at a resolution of \(640 \times 480\) pixels, while the original depth images had a resolution of \(1280 \times 800\) pixels. After image alignment, the depth images were resized to match the resolution of the RGB images. Our dataset represents a unique effort to collect RGB-D images with a specific two-view setup derived from an actual robotic platform used in commercial products developed by Fujian Hantewin Intelligent Technology Co., Ltd. See Tab.~\ref{tab1} for a comparison of different glass-related RGB-D datasets.

\begin{table*}[t!]
\renewcommand{\arraystretch}{1.25}
\centering
\caption{Comparison of different glass-related RGB-D datasets, including the number of images, types of environments covered, and class distributions.}
\label{tab:dataset_compare}
\begin{tabular}{l p{2.2cm} p{6.3cm} p{3cm}} 
\toprule
\textbf{Dataset} & \textbf{Number of Images} & \multicolumn{1}{c}{\textbf{Types of Environments Covered}} & \textbf{Class Distributions} \\
\midrule
MJU-Glass   & 4,900   & Indoor and outdoor scenes, including extreme environments & Glass, background \\
RGBD-Mirror & 3,049   & Various indoor environments with glasses                  & Glass, background \\
GW-Depth    & 1,200   & Indoor and outdoor glass wall scenes                      & Glass walls, background \\
\bottomrule
\end{tabular}
\label{tab1}
\end{table*}

\begin{figure}[htbp]
\centerline{\includegraphics[width=1\linewidth]{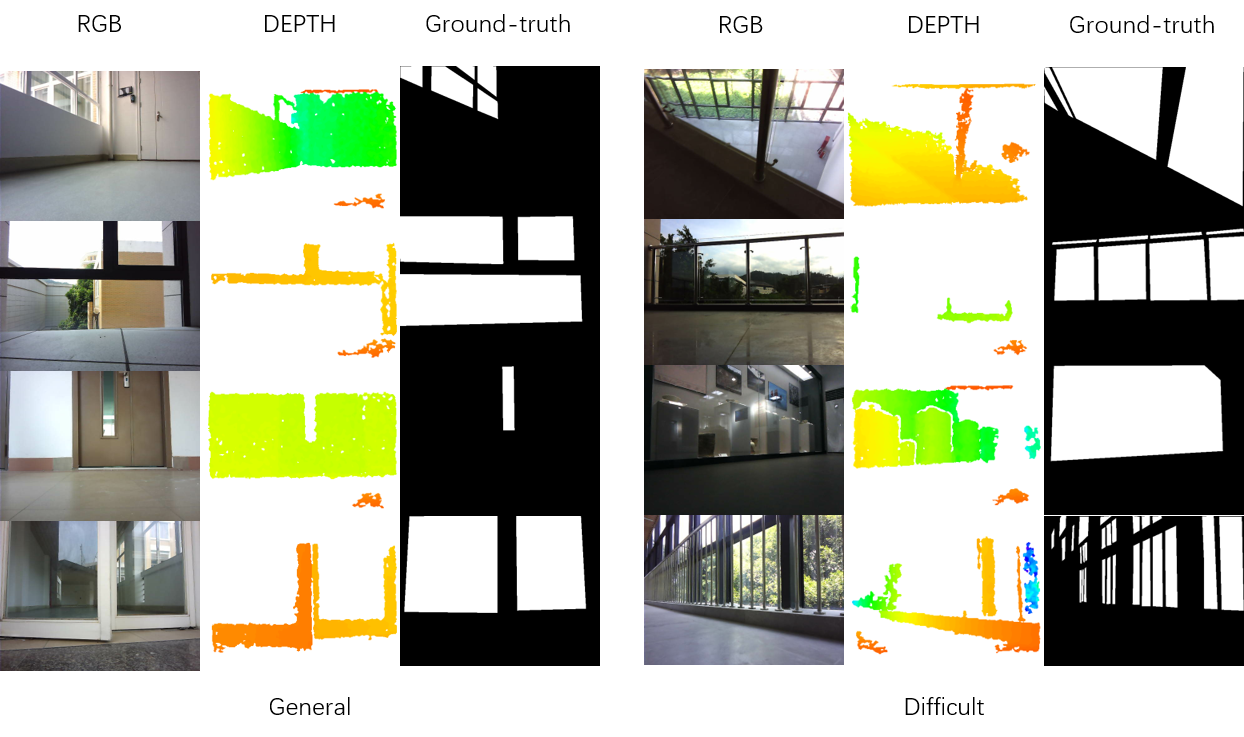}}
\caption{Examples RGB-D images from the MJU-Glass dataset, highlighting the diversity of scenes and challenging conditions encountered. The left column shows general test images, while the right column shows images from the difficult test set.}
\label{fig4}
\end{figure}

\textbf{Difficult Test Set}: Since the model focuses on a binary pixelwise classification task (glass and background), we sorted the test results of widely used models and assembled a subset of 400 images with the lowest mIoU from DeepLabv3+ and PSPNet into a difficult test dataset. These challenging scenarios include bright lighting, high transparency, and complex environments, which are critical for robotic applications. Example images from the general test set and the difficult test set are shown in Fig.~\ref{fig4}. Some of the typical challenging scenarios in the difficult test set include: (1) bright lighting, which causes the degradation of the depth images, (2) poor lighting, which causes the appearance features to be less effective, (3) high transparency, which implies that objects behind the glass can be seen more clearly, therefore glass surfaces will be more difficult to recognize, and (4) complex environments, which imply that the scene is complicated by the superimposition of multiple objects. The difficult test set aims to evaluate the robustness and performance of the model under extreme conditions, providing valuable insights into its strengths and limitations.
\begin{figure}[htbp]
\centerline{\includegraphics[width=1\linewidth]{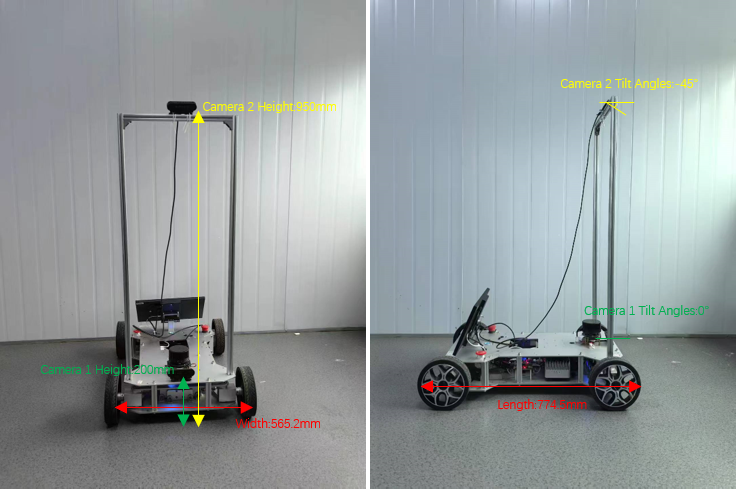}}
\caption{Front and side views of the mobile robot platform we used to collect the MJU-Glass dataset. Two Gemini Pro RGB-D cameras from Orbbec are mounted at different heights and angles to cover a wider view.}
\label{fig5}
\end{figure}

In order to reproduce the perspective of service robots in their daily activities, we assembled a mobile robot with two cameras of different heights and angles, as shown in Fig.~\ref{fig5}. The mobile robot used for dataset acquisition has a compact design with dimensions of \(565.2 mm\times774.5 mm\times180 mm\) and a maximum speed of \(0.46 m/s\). It is equipped with an NVIDIA Jetson TX1 as the onboard computer, providing sufficient computational power for real-time image processing and control tasks. Two RGB-D cameras are mounted on the robot at different heights and angles to enable multi-view perception. Camera 1 is installed at a height of \(200 mm\) forming a \(0^\circ\) angle with the horizontal ground. This configuration is suitable for capturing close-range or ground-level data. Camera 2 is positioned at a height of 950mm with a tilt angle of \(-{45}^\circ\) relative to the horizon, allowing it to observe the ground and close-up objects with a larger projected size, making it suitable for tasks such as object detection and spatial mapping. In terms of specifications, the  depth sensing module has a working range of \(0.25 m\) to \(2.5 m\) with a precision of \(\pm 5 mm\) at~\(1 m\). The depth images are captured at a resolution of \(640\times400\)pixels and a frame rate of 60fps, with a field of view (FOV) of \({67.9}^\circ\left(H\right)\times{45.3}^\circ\left(V\right)\). The RGB module captures images at a resolution of \(640\times480\)pixels and the same frame rate of 60fps, with a wider FOV of \({71.5}^\circ\left(H\right)\times{56.7}^\circ\left(V\right)\). These specifications indicate that the cameras provide high frame rate and precision along with a wide FOV, making them well-suited for dynamic scene capture and 3D reconstruction tasks. 

\subsection{Experimental Settings}
\textbf{Model Selection}: In our experiments, we choose DeepLabv3+~\cite{Chen2018EncoderDecoderWA} and PSPNet~\cite{Zhao2016PyramidSP} as benchmark networks for evaluating the proposed WFF module. In addition, a ResNet-50 backbone with ImageNet pretrained weights was used, which helps to accelerate convergence and improve performance in the glass surface segmentation task.

Our goal is to distinguish between glass and background, so our problem is defined as a binary pixelwise segmentation task. The following quantitative performance metrics are adopted: the Intersection over Union (IoU) for glass, the mean IoU (mIoU) for glass and background combined, and the boundary IoU (bIoU). Let TP, FP, and FN represent the total number of true positive, false positive, and false negative pixels, respectively. The three evaluation criteria are defined as follows:

\textbf{IoU}: Intersection over Union (IoU) represents the overlap between the predicted segmentation mask and the ground truth, reflecting the ability of the model to predict specific objects. In the following, the IoU targets are glass, excluding the background.
\begin{equation}
IoU_{Glass}=\frac{TP_{Glass}}{TP_{Glass}+FP_{Glass}+FN_{Glass}},
\end{equation}

\noindent Here, \(TP_{Glass}\), \(FP_{Glass}\), and \(FN_{Glass}\) represent the true positives, false positives, and true negatives, respectively, specifically for the glass class. Since there is only one foreground class, we use IoU to denote \(IoU_{Glass}\) from now on. These values are also included in the mIoU below, i.e., \(TP_c\), \(FP_c\), and \(FN_c\), where c is the class (glass or background).

\textbf{mIoU}: Mean IoU (mIoU) is calculated by averaging the IoU values for both the background and glass classes, reflecting the overall prediction performance of the model. In the following, $c$ represents the class, where $0$ denotes background and $1$ denotes glass. The calculation is summarized in Eqn.~\ref{eqn9}.
\begin{equation}
mIoU=\frac{1}{2}\sum_{c=0}^{1}\frac{TP_c}{TP_c+FP_c+FN_c}
\label{eqn9}
\end{equation}

\textbf{bIoU}: Boundary IoU (bIoU) measures the intersection between the segmented boundary of the target and the ground-truth boundary within a $5$-pixel range, reflecting the capability of the model in handling boundary details. The calculation is summarized in Eqn.~\ref{eqn10}.
\begin{equation}
bIoU=\frac{TP_{boundary}}{TP_{boundary}+FP_{boundary}+FN_{boundary}}
\label{eqn10}
\end{equation}

\subsection{Results}
In this section, we demonstrate the effectiveness of the module through two sets of experiments, a comparison experiment and an ablation experiment.

In Tab.~\ref{tab2}, we can clearly observe that with the incorporation of the WFF module, both DeepLabv3+ and PSPNet achieve the best performance across all evaluation metrics on both the All and Difficult test datasets. The competing methods are both the baseline approach with RGB images only, and a simple RGB-D feature fusion method with concatenation. It is worth noting that mIoU tends to show less noticeable gains partly because it includes the background—which is typically the largest area and easier to segment accurately—while the boundary IoU (bIoU) metric, reflecting performance on object boundaries, is much more challenging. The observed improvements in bIoU, especially on the difficult test dataset where boundary details are critical, indicate that the WFF module is effective in enhancing the model’s capability to delineate complex and fine structures, thereby addressing one of the most challenging aspects of segmentation.

\begin{table*}
\renewcommand{\arraystretch}{1.2}
\caption{Performance comparison on the MJU-Glass dataset using RGB-only models, simple feature fusion, and the proposed WFF module.}
\label{tab:exp_compare}
\centering
\begin{tabular}{llccc|ccc}
\toprule
\textbf{Method} & \textbf{Backbone} & \multicolumn{3}{c|}{\textbf{All}} & \multicolumn{3}{c}{\textbf{Difficult}} \\
 & & \textbf{IoU} & \textbf{mIoU} & \textbf{bIoU} & \textbf{IoU} & \textbf{mIoU} & \textbf{bIoU} \\
\midrule
\multicolumn{8}{l}{\textit{Baseline (RGB only)}} \\
DeepLabv3+       & ResNet-50 & 89.61 & 93.17 & 38.09 & 88.74 & 92.12 & \textbf{38.10} \\
PSPNet           & ResNet-50 & 86.74 & 91.43 & 25.33 & 85.90 & 90.31 & \textbf{25.60} \\
\midrule
\multicolumn{8}{l}{\textit{Simple Feature Fusion (Concatenation)}} \\
DeepLabv3+-concat & ResNet-50 & 89.76 & 93.36 & 39.91 & 88.74 & 92.21 & \textbf{38.56} \\
PSPNet-concat     & ResNet-50 & 90.21 & 93.67 & 32.66 & 88.74 & 91.49 & \textbf{30.17} \\
\midrule
\multicolumn{8}{l}{\textit{Proposed Weighted Feature Fusion (WFF)}} \\
DeepLabv3+-WFF    & ResNet-50 & 90.12 & 93.61 & 40.35 & 88.80 & 92.36 & \textbf{39.44} \\
PSPNet-WFF        & ResNet-50 & \textbf{90.47} & \textbf{93.84} & \textbf{33.69} & \textbf{89.39} & \textbf{92.43} & \textbf{33.09} \\
\bottomrule
\end{tabular}
\label{tab2}
\end{table*}

In the ablation experiments (Tab.~\ref{tab3}), \(F_{af}\) and \(F_{aw}\) represent different components of the WFF module, specifically feature summation and weight adjustment, respectively. The addition of \(F_{af}\), which facilitates feature summation, improves the ability of the model to capture richer and more integrated information from different inputs, leading to better overall segmentation performance. On the other hand, \(F_{aw}\), which adjusts the weight of features automatically, plays a more crucial role in refining the boundaries of the segmented objects. This is particularly important for the accurate delineation of the glass region in the presence of complex background details. As a result, \(F_{aw}\) contributes more significantly to boundary refinement, while both features together improve the overall accuracy, as shown by the consistent performance improvement in the experiments. We also notice that the automatic weight adjustment has a greater impact on bIoU, this is because there is a lot of noise at the edges in the depth map, and simple fusion cannot highlight the features of the edges, while the automatic weight adjustment can be better handle the rich contextual information in the RGB images and to combine the depth information in the depth images.
\begin{table*}
\centering
\caption{Ablation experiments to validate the efficacy of different components in the WFF module.}
\begin{tabular}{ccclll}
\hline
\multicolumn{6}{l}{\textbf{DeepLabv3+ with MJU-Glass}}                                                                                                                                          \\ \hline
\multicolumn{1}{l}{\textbf{Baseline}} & \multicolumn{1}{l}{\textbf{$\boldsymbol{F_{af}}$}} & \multicolumn{1}{l|}{\textbf{$\boldsymbol{F_{aw}}$}} & \textbf{IoU} & \textbf{mIoU} & \textbf{bIoU} \\ \hline
\checkmark                            &                                                    & \multicolumn{1}{c|}{}                               & 89.61        & 93.17        & 38.09         \\
\checkmark                            & \checkmark                                         & \multicolumn{1}{c|}{}                               & 89.76        & 93.36        & 39.91        \\
\checkmark                            & \checkmark                                         & \multicolumn{1}{c|}{\checkmark}                     & 90.12        & 93.61         & 40.35        \\ \hline
\multicolumn{6}{l}{}                                                                                                                                                                            \\ \hline
\multicolumn{6}{l}{\textbf{PSPNet with MJU-Glass}}                                                                                                                                              \\ \hline
\multicolumn{1}{l}{\textbf{Baseline}} & \multicolumn{1}{l}{\textbf{$\boldsymbol{F_{af}}$}} & \multicolumn{1}{l|}{\textbf{$\boldsymbol{F_{aw}}$}} & \textbf{IoU} & \textbf{mIoU} & \textbf{bIoU} \\ \hline
\checkmark                            &                                                    & \multicolumn{1}{c|}{}                               & 86.74       & 91.43       & 25.33       \\
\checkmark                            & \checkmark                                         & \multicolumn{1}{c|}{}                               & 90.21\        & 93.67       & 33.60         \\
\checkmark                            & \checkmark                                         & \multicolumn{1}{c|}{\checkmark}                     & 90.47       & 93.84        & 33.69       \\ \hline
\end{tabular}
\label{tab3}
\end{table*}

In the WFF module, the main focus is on dynamically adjusting the contribution of RGB and depth image features to the final feature map by performing weighted fusion on them. The weighted fusion process involves processing and combining RGB and depth features and normalizing the weights using softmax. These operations are mainly implemented through the fully-connected and pooling layers, and are all relatively computationally affordable operations. Throughout the process, feature fusion is essentially performed through simple addition operations (summing RGB and depth features), followed by processing and weighting through a small number of fully connected layers. Therefore, although the WFF module adds a small computational overhead, these operations do not significantly increase the computational complexity, when compared to the original model. The design of the WFF module focuses on guaranteeing the accuracy of the model by optimizing the weight adjustment and feature fusion, without having a large negative impact on the inference speed (frames per second).

In conventional RGB-only recognition models, the special properties of glass, such as high transparency and strong reflectivity, make the environment more complex and difficult to recognize, posing significant challenges to glass surface segmentation. The advent of depth cameras has addressed the difficulty of recognizing glass due to its distinctive missing depth patterns. Based on this premise, by combining the RGB and depth images, the different features from both modalities can better localize glass regions.
\begin{figure}[htbp]
\centerline{\includegraphics[width=1\linewidth]{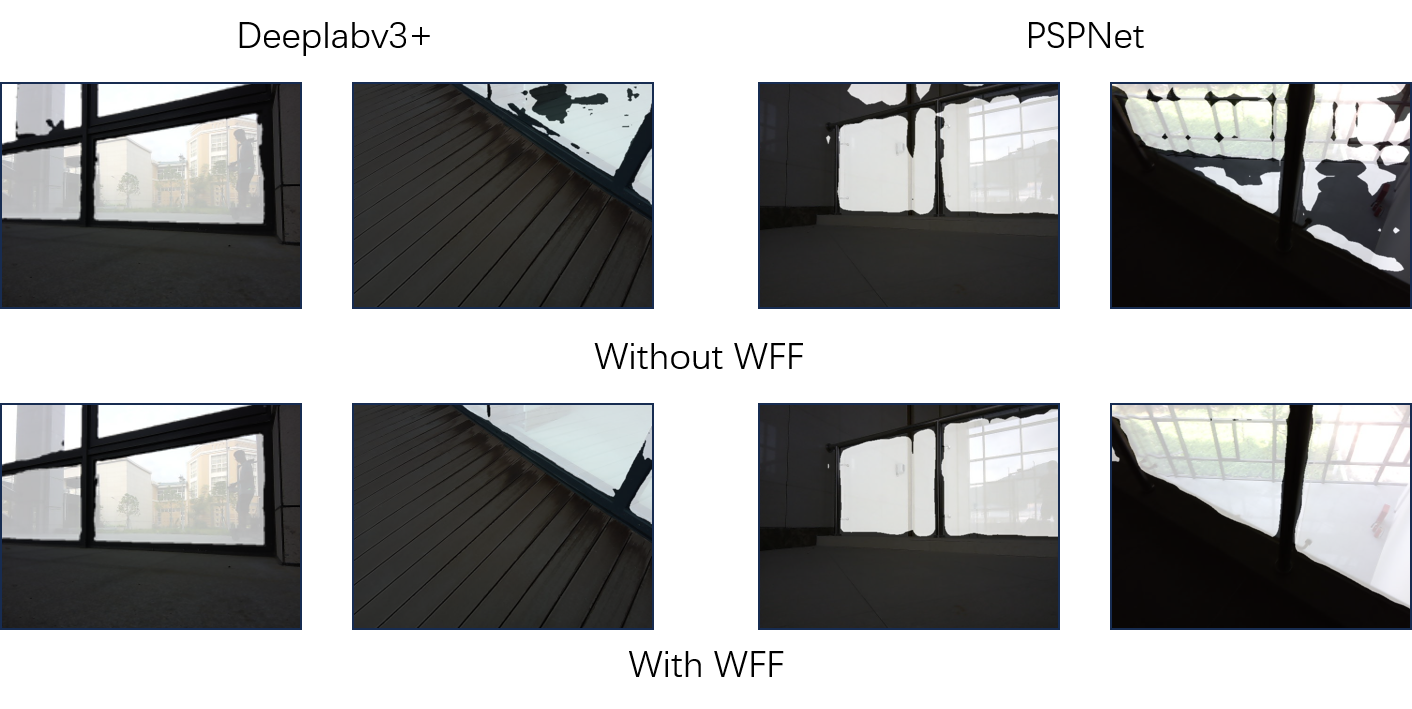}}
\caption{Example RGB-D glass surface segmentation results using DeepLabv3+ and PSPNet with and without the WFF module.}
\label{fig6}
\end{figure}

As we can see from the qualitative segmentation examples in Fig.~\ref{fig6}, with the addition of the WFF module, the model is able to better handle reflections, highlights, and interference from objects behind the glass. Specifically, the WFF module enhances the ability of the model to follow glass boundaries while reducing the impact of background clutter, improving segmentation accuracy in challenging environments.

\section{CONCLUSION}
In this work, we proposed the Weighted Feature Fusion (WFF) module and introduced the MJU-Glass dataset, designed for glass surface segmentation. Our proposed method and experiments focus on service robots, where accurate segmentation of transparent surfaces is vital for tasks like obstacle avoidance and navigation in environments with reflections. Experiments show that the WFF module effectively balances the contribution of RGB and depth features, improving robustness and accuracy in challenging scenarios involving transparent objects. In addition, the MJU-Glass dataset serves as a new public benchmark for glass surface segmentation models. Future work could integrate additional modalities like thermal or polarization data to enhance segmentation accuracy. The MJU-Glass dataset could also be extended to other computer vision tasks, such as object detection in transparent environments, further broadening its applications. Additionally, exploring transformer-based architectures and alternative loss functions could lead to further performance improvements.

\bibliographystyle{IEEEtran}  
\bibliography{references}

\begin{thebibliography}{10}
\providecommand{\url}[1]{#1}
\csname url@samestyle\endcsname
\providecommand{\newblock}{\relax}
\providecommand{\bibinfo}[2]{#2}
\providecommand{\BIBentrySTDinterwordspacing}{\spaceskip=0pt\relax}
\providecommand{\BIBentryALTinterwordstretchfactor}{4}
\providecommand{\BIBentryALTinterwordspacing}{\spaceskip=\fontdimen2\font plus
\BIBentryALTinterwordstretchfactor\fontdimen3\font minus
  \fontdimen4\font\relax}
\providecommand{\BIBforeignlanguage}[2]{{%
\expandafter\ifx\csname l@#1\endcsname\relax
\typeout{** WARNING: IEEEtran.bst: No hyphenation pattern has been}%
\typeout{** loaded for the language `#1'. Using the pattern for}%
\typeout{** the default language instead.}%
\else
\language=\csname l@#1\endcsname
\fi
#2}}
\providecommand{\BIBdecl}{\relax}
\BIBdecl

\bibitem{Ronneberger2015UNetCN}
\BIBentryALTinterwordspacing
O.~Ronneberger, P.~Fischer, and T.~Brox, ``U-net: Convolutional networks for
  biomedical image segmentation,'' \emph{ArXiv}, vol. abs/1505.04597, 2015.
  [Online]. Available: \url{https://api.semanticscholar.org/CorpusID:3719281}
\BIBentrySTDinterwordspacing

\bibitem{Zhao2016PyramidSP}
\BIBentryALTinterwordspacing
H.~Zhao, J.~Shi, X.~Qi, X.~Wang, and J.~Jia, ``Pyramid scene parsing network,''
  \emph{2017 IEEE Conference on Computer Vision and Pattern Recognition
  (CVPR)}, pp. 6230--6239, 2016. [Online]. Available:
  \url{https://api.semanticscholar.org/CorpusID:5299559}
\BIBentrySTDinterwordspacing

\bibitem{Zhang2022SegViTSS}
\BIBentryALTinterwordspacing
B.~Zhang, Z.~Tian, Q.~Tang, X.~Chu, X.~Wei, C.~Shen, and Y.~Liu, ``Segvit:
  Semantic segmentation with plain vision transformers,'' \emph{ArXiv}, vol.
  abs/2210.05844, 2022. [Online]. Available:
  \url{https://api.semanticscholar.org/CorpusID:252846611}
\BIBentrySTDinterwordspacing

\bibitem{Shelhamer2014FullyCN}
\BIBentryALTinterwordspacing
E.~Shelhamer, J.~Long, and T.~Darrell, ``Fully convolutional networks for
  semantic segmentation,'' \emph{2015 IEEE Conference on Computer Vision and
  Pattern Recognition (CVPR)}, pp. 3431--3440, 2014. [Online]. Available:
  \url{https://api.semanticscholar.org/CorpusID:1629541}
\BIBentrySTDinterwordspacing

\bibitem{Chen2017RethinkingAC}
\BIBentryALTinterwordspacing
L.-C. Chen, G.~Papandreou, F.~Schroff, and H.~Adam, ``Rethinking atrous
  convolution for semantic image segmentation,'' \emph{ArXiv}, vol.
  abs/1706.05587, 2017. [Online]. Available:
  \url{https://api.semanticscholar.org/CorpusID:22655199}
\BIBentrySTDinterwordspacing

\bibitem{Zheng2020RethinkingSS}
\BIBentryALTinterwordspacing
S.~Zheng, J.~Lu, H.~Zhao, X.~Zhu, Z.~Luo, Y.~Wang, Y.~Fu, J.~Feng, T.~Xiang,
  P.~H.~S. Torr, and L.~Zhang, ``Rethinking semantic segmentation from a
  sequence-to-sequence perspective with transformers,'' \emph{2021 IEEE/CVF
  Conference on Computer Vision and Pattern Recognition (CVPR)}, pp.
  6877--6886, 2020. [Online]. Available:
  \url{https://api.semanticscholar.org/CorpusID:229924195}
\BIBentrySTDinterwordspacing

\bibitem{Huo2022GlassSW}
\BIBentryALTinterwordspacing
D.~Huo, J.~Wang, Y.~Qian, and Y.-H. Yang, ``Glass segmentation with rgb-thermal
  image pairs,'' \emph{IEEE Transactions on Image Processing}, vol.~32, pp.
  1911--1926, 2022. [Online]. Available:
  \url{https://api.semanticscholar.org/CorpusID:248118556}
\BIBentrySTDinterwordspacing

\bibitem{Mei2022GlassSU}
\BIBentryALTinterwordspacing
H.~Mei, B.~Dong, W.~Dong, J.~Yang, S.-H. Baek, F.~Heide, P.~Peers, X.~Wei, and
  X.~Yang, ``Glass segmentation using intensity and spectral polarization
  cues,'' \emph{2022 IEEE/CVF Conference on Computer Vision and Pattern
  Recognition (CVPR)}, pp. 12\,612--12\,621, 2022. [Online]. Available:
  \url{https://api.semanticscholar.org/CorpusID:249916399}
\BIBentrySTDinterwordspacing

\bibitem{Wang2012GlassOL}
\BIBentryALTinterwordspacing
T.~Wang, X.~He, and N.~Barnes, ``Glass object localization by joint inference
  of boundary and depth,'' \emph{Proceedings of the 21st International
  Conference on Pattern Recognition (ICPR2012)}, pp. 3783--3786, 2012.
  [Online]. Available: \url{https://api.semanticscholar.org/CorpusID:18189708}
\BIBentrySTDinterwordspacing

\bibitem{Huang2018GlassDA}
\BIBentryALTinterwordspacing
Z.~Huang, K.~Wang, K.~Yang, R.~Cheng, and J.~Bai, ``Glass detection and
  recognition based on the fusion of ultrasonic sensor and rgb-d sensor for the
  visually impaired,'' in \emph{Security + Defence}, 2018. [Online]. Available:
  \url{https://api.semanticscholar.org/CorpusID:70281006}
\BIBentrySTDinterwordspacing

\bibitem{Zhao2023GlassDI}
\BIBentryALTinterwordspacing
Y.~Zhao, H.~Li, S.~Jiang, H.~Li, Z.~Zhang, and H.~Zhu, ``Glass detection in
  simultaneous localization and mapping of mobile robot based on rgb-d
  camera,'' \emph{2023 IEEE International Conference on Mechatronics and
  Automation (ICMA)}, pp. 549--556, 2023. [Online]. Available:
  \url{https://api.semanticscholar.org/CorpusID:261107431}
\BIBentrySTDinterwordspacing

\bibitem{zhu2024mffnet}
L.~Zhu, T.~Li, Y.~Ning, and Y.~Zhang, ``Mffnet: Multimodal feature fusion
  network for rgb-d transparent object detection,'' \emph{International Journal
  of Advanced Robotic Systems}, vol.~21, no.~5, p. 17298806241283373, 2024.

\bibitem{lin2025leveraging}
J.~Lin, Y.-H. Yeung, S.~Ye, and R.~W. Lau, ``Leveraging rgb-d data with
  cross-modal context mining for glass surface detection,'' in
  \emph{Proceedings of the AAAI Conference on Artificial Intelligence},
  vol.~39, no.~5, 2025, pp. 5254--5261.

\bibitem{He2015DeepRL}
\BIBentryALTinterwordspacing
K.~He, X.~Zhang, S.~Ren, and J.~Sun, ``Deep residual learning for image
  recognition,'' \emph{2016 IEEE Conference on Computer Vision and Pattern
  Recognition (CVPR)}, pp. 770--778, 2015. [Online]. Available:
  \url{https://api.semanticscholar.org/CorpusID:206594692}
\BIBentrySTDinterwordspacing

\bibitem{Mei2021DepthAwareMS}
\BIBentryALTinterwordspacing
H.~Mei, B.~Dong, W.~Dong, P.~Peers, X.~Yang, Q.~Zhang, and X.~Wei,
  ``Depth-aware mirror segmentation,'' \emph{2021 IEEE/CVF Conference on
  Computer Vision and Pattern Recognition (CVPR)}, pp. 3043--3052, 2021.
  [Online]. Available: \url{https://api.semanticscholar.org/CorpusID:235357151}
\BIBentrySTDinterwordspacing

\bibitem{Chang2017Matterport3DLF}
\BIBentryALTinterwordspacing
A.~X. Chang, A.~Dai, T.~A. Funkhouser, M.~Halber, M.~Nie{\ss}ner, M.~Savva,
  S.~Song, A.~Zeng, and Y.~Zhang, ``Matterport3d: Learning from rgb-d data in
  indoor environments,'' \emph{2017 International Conference on 3D Vision
  (3DV)}, pp. 667--676, 2017. [Online]. Available:
  \url{https://api.semanticscholar.org/CorpusID:21435690}
\BIBentrySTDinterwordspacing

\bibitem{Song2015SUNRA}
\BIBentryALTinterwordspacing
S.~Song, S.~P. Lichtenberg, and J.~Xiao, ``Sun rgb-d: A rgb-d scene
  understanding benchmark suite,'' \emph{2015 IEEE Conference on Computer
  Vision and Pattern Recognition (CVPR)}, pp. 567--576, 2015. [Online].
  Available: \url{https://api.semanticscholar.org/CorpusID:6242669}
\BIBentrySTDinterwordspacing

\bibitem{Fan2019RethinkingRS}
\BIBentryALTinterwordspacing
D.-P. Fan, Z.~Lin, J.~Zhao, Y.~Liu, Z.~Zhang, Q.~Hou, M.~Zhu, and M.-M. Cheng,
  ``Rethinking rgb-d salient object detection: Models, data sets, and
  large-scale benchmarks,'' \emph{IEEE Transactions on Neural Networks and
  Learning Systems}, vol.~32, pp. 2075--2089, 2019. [Online]. Available:
  \url{https://api.semanticscholar.org/CorpusID:196831465}
\BIBentrySTDinterwordspacing

\bibitem{Armeni2017Joint2D}
\BIBentryALTinterwordspacing
I.~Armeni, S.~Sax, A.~Zamir, and S.~Savarese, ``Joint 2d-3d-semantic data for
  indoor scene understanding,'' \emph{ArXiv}, vol. abs/1702.01105, 2017.
  [Online]. Available: \url{https://api.semanticscholar.org/CorpusID:2730848}
\BIBentrySTDinterwordspacing

\bibitem{Liang2023MonocularDE}
\BIBentryALTinterwordspacing
Y.~Liang, B.~Deng, W.~Liu, J.~Qin, and S.~He, ``Monocular depth estimation for
  glass walls with context: A new dataset and method,'' \emph{IEEE Transactions
  on Pattern Analysis and Machine Intelligence}, vol.~45, pp. 15\,081--15\,097,
  2023. [Online]. Available:
  \url{https://api.semanticscholar.org/CorpusID:261174309}
\BIBentrySTDinterwordspacing

\bibitem{Chen2018EncoderDecoderWA}
\BIBentryALTinterwordspacing
L.-C. Chen, Y.~Zhu, G.~Papandreou, F.~Schroff, and H.~Adam, ``Encoder-decoder
  with atrous separable convolution for semantic image segmentation,'' in
  \emph{European Conference on Computer Vision}, 2018. [Online]. Available:
  \url{https://api.semanticscholar.org/CorpusID:3638670}
\BIBentrySTDinterwordspacing

\end{thebibliography}

\end{document}